\documentclass[english,utf8x]{article-hermes}

\journal{TAL. Volume 55 -- n° 3/2014}{7}{19}

\title[Introduction]{Introduction: Cognitive Issues in Natural Language Processing}

\author{Thierry Poibeau\fup{*} \andauthor Shravan Vasishth\fup{**}  } 

\address{%
\fup{*} Laboratoire LATTICE \andauthor CNRS, \'Ecole normale sup\'erieure and Universit\'e Sorbonne Nouvelle, France\\
\fup{**} University of Potsdam, Germany\\
}
\begin{document}

\maketitlepage

\section{Introduction}

From the very beginning, pioneers in computer sciences have tried to establish a link between practical issues and more fundamental ones, like language processing in the brain, knowledge representation or  the nature of communication. Modeling language was supposed to open a window on the brain, or at least give an idea of how things can be processed by the brain. The famous Turing Test \cite{turing1950,peters2004} itself was not directly about language modeling, but mimicking a conversation through a computer that was considered as a proof of intelligence, i.e. the fact that computers could interpret utterances, infer new information and produce relevant responses. 

These concerns never disappeared but the field quickly moved towards more applied research interests. It was especially the case for machine translation, as soon as in the 1950s, since it was important then to show that usable results could be obtained rapidly \cite{hutchins1986,hutchins2000}. However, there has been a continuous interest since the Second World War for what could be called ``cognitive issues''. Even the deception about the poor performances of machine translation systems in the 1960s led to reflections about language complexity and whether it was possible to model this complexity or not. In the 1960s and 1970s, Robert Dreyfus' works  have for example been largely influenced by the results of programmes dealing with computers and languages \cite{dreyfus1965,dreyfus1972}. 

Dreyfus advocates, among other things, that a large part of language production is unconscious or, at least, cannot be modeled via explicit rule-based systems. Neural networks have then been seen as a possible solution to model unconscious reasoning: the architecture of natural language processing was  supposed to offer a more sensible representation of the problem \cite{rumelhart1986}. It is now generally assumed that neural networks do not directly simulate the human brain but they allow to implement processes that are required to model language and other complex tasks. 

It seems that this interest for cognitive issues decreased in the late 1980s and more dramatically during the 1990s. The need for practical and real world applications pushed forward more applied and engineering approaches, with a decreasing interest for cognitive issues, even if of course some researchers have always been active in this field. One sign of this situation is for example that even if recent conferences like the Conference on Natural Language Learning (CoNLL) or Empirical Methods in Natural Language Processing (EMNLP) always had a track on cognitive and psycholinguistic issues, very few submissions were generally received and even less accepted\footnote{One should however note that more established conferences like ACL or COLING never had a cognitive track \textit{per se}, even in the 1960s or 1970s. However, each year several papers were dealing with issues related to cognitive science, like the structure of the mental lexicon or the psychological plausibility of parsing algorithms. }. The two domains have then largely evolved separately, cognitive science researchers being maybe overwhelmed by the high technicality of NLP nowadays, while natural language processing researchers do not always see the added value of taking cognitive aspects into account for their problems. 

However, things seem to be changing. Several workshops are now regularly held on these topics (see for example the ``Cognitive Aspects of Computational Language Learning'' workshop regularly held since 2007 \cite{villavicencio13} or the ``Cognitive Modeling and Computational Linguistics''  workshop during NAACL 2015). Natural language processing would benefit from a better understanding of human processes: as it has been said by several researchers recently, traditional machine learning approaches have brought rapid and important improvements in different natural language processing tasks but these successes may be ``low hanging fruits'' \cite{church2011}, which means the field may have to face from now on more difficult problems  (e.g. discourse planning, argumentative analysis, etc.) that would benefit from a better understanding of the processes involved in the brain. 

Additionally, researchers are again interested in evaluating the relevance of their models according to a cognitive dimension. Many of the existing computational models attempt to study language tasks under cognitively plausible criteria (such as memory and processing limitations)
that humans face \cite{blache13}. New machine learning techniques, especially deep learning, bring back to the front scene a new version of neural networks that seems both more powerful and more sound, from a technical as well as a cognitive point of view \cite{socher2013}. Last but not least, cognitive science also benefit and sometimes takes inspiration from computational models.

\section{Current research trends} 

In this section, we examine some recent research trends concerning language comprehension, language acquisition, language pathologies and language evolution. There are of course other domains where computational linguistics meets cognitive science but we think that these four areas have seen important improvements recently. 

\subsection{Language comprehension}

Cognitive science has a strong tradition of developing models of cognitive processes using tools that originated in the natural language processing world. One of the early examples in sentence comprehension research is the work by Joshi and colleagues \cite{joshi90,rambow94processing}, who developed an explanation of the differences in processing between crossed and nested dependencies in Dutch and German \cite{bachetal86} by explicitly linking a type of pushdown automaton, the Bottom-up Embedded Pushdown Automaton (BEPDA), to Tree Adjoining Grammar (TAG) \cite{joshi00}.  A unique aspect of their work was that they did not directly use TAG to develop a processing model, but defined an equivalence relationship between the automaton and TAG grammars.  


Over the last fifteen years, an information-theoretic approach to language comprehension processes, led by \cite{haleearley01,hale03,hale06} and Roger Levy \cite{levy08}, has generated a lot of interest, especially because it was able to explain certain empirical findings  \cite{lars00,vasishthlewisLanguage05} that needed additional assumptions to account for using classical working memory-based accounts \cite{just1980tre,gibson00}. 
The Hale and Levy approach relied on large-scale probabilistic context free grammars  derived from treebanks, and consists of computing a complexity metric, such as surprisal or entropy reduction, to characterize processing difficulty. The distinguishing feature of this body of work is that it formalizes the effect that the frequency of grammatical continuations has on expectations about upcoming words and phrases. The surprisal idea had already been investigated decades earlier, and has a rich tradition in EEG research \cite{KutasHillyard1984}, but the Hale and Levy approach provided a formal basis for computing surprisal on a moment-by-moment basis. The Levy paper directly led to attempts to experimentally evaluate the predictions of the surprisal and entropy reduction ideas using large-scale grammars and eyetracking corpora or other reading data \cite{jemrsurprisal,demberg2008data,bostonhalevasishthklieglLCP09,frank2013uncertainty}. In addition, planned experiments also tested specific predictions of the surprisal account, with mixed results.  Several studies have found evidence largely consistent with surprisal and related metrics \cite{staub2010eye,levy2013expectation,FrankEtAl2015,Jaegeretal2015},
but other research has shown mixed evidence consistent with both the surprisal and the classical working memory-based accounts  \cite{levy2013syntactic,HusainEtAl2014,HusainVasishthNarayanan2015,SafaviEtAlCUNY2015}. As Levy pointed out in his 2008 paper, it is likely that both classes of explanation play a role in determining comprehension difficulty. Nevertheless, a major contribution of the probabilistic grammar-based approach has  been to give a formal foundation to the idea of expectation driving parsing processes.
Future work would profit from building probabilistic parsing frameworks that include both kinds of constraints; this has been attempted in the past \cite{bostonhalevasishthklieglLCP09}, but the field would greatly benefit from more principled model development and model evaluation over cross-linguistic data. A cross-linguistic investigation is necessary as a corrective to the tendency to focus on English as the target language. A further gap in the field is the apparent lack of connection between the Hale and Levy formalization of expectation, and the rich body of work on surprisal in the EEG literature. Testing the detailed predictions of surprisal and related ideas using EEG is an important empirical test that is yet to be conducted. Another area where proposals such as surprisal and entropy reduction can make important contributions is individual-level variability in expectations, arising from differences in working memory capacity and the varying grammatical knowledge of the comprehender \cite{NicenboimEtAl2015article}. 

Ten years ago, it seemed that two areas where probabilistic models needed most work was to (i) demonstrate that moment-by-moment processing can be explained in terms of probabilistic expectation, and to (ii) attempt to model a broader range of phenomena. In the last few years, excellent progress has been made regarding the first point, but the empirical coverage of the probabilistic parsing account remains rather limited. For example, there is a large literature---and a great deal of empirical evidence---on interference effects arising in agreement attraction configurations \cite{wagersetal}, and antecedent-reflexive constructions where principles of the binding theory play a role \cite{Dillon-EtAl-2013}. Probabilistic models obviously have little to say about phenomena that involve linguistic as opposed to probabilistic constraints. This highlights the need for developing more comprehensive models of cognitive processes such as sentence comprehension that take a broader view of the constraints that act on cognition.

\subsection{Language acquisition}

Language acquisition is another domain where computation models and cognitive sciences have a fruitful dialogue. To a certain extent, neuro-imaging has renewed the study of language by making it possible to directly observe processes in the brain \cite{dehaene07} but for the rest, language is only known through direct production, i.e. language utterances. Therefore, the study of language acquisition by children is crucial, since it gives an overview on what vocabulary and structures are mastered first, what untypical constructions (when compared to adult speech) are used by children during learning, etc. 

The study of language acquisition has seen important developments recently thanks to the development of automatic approaches. The main reason for these new approaches is the availability of large amont of data in the Childes database \cite{macwhinney95}, especially for English, but also for other languages, French being especially well represented here \cite{morgenstern12}. Most recordings concern discussions between children and adults in real life situations (during dinner, bath, play, etc.). Videos are generally available so as to allow a multimodal approach to the study of language acquisition. Most corpora are annotated with morphosyntactic information \cite{sagae07}, and parts of the corpus have received other kinds of annotation, including syntactic tags \cite{sagae10} and more rarely semantic or multimodal annotations (concerning gestures for example) \cite{morgenstern12}. The Talkbank project has accumulated multimodal information in a more systematic way \cite{macwhinney04}. One should also mention Roy, who has systematically recorded the first years of his son, producing thousands of hours of videos \cite{roy09}. However, this mass of data is not structured and not tagged, making it hard to be used by researchers, on top of the ethic problem that this kind of recordings may pose. One specificity of most of these corpora is to provide longitudinal data. They generally make it possible to follow language development of a single child over several months. 

Language acquisition studies concern mainly the development of the language ability in young children, the complexity of their speech in accordance with their age, and the comparison of children productions with adult ones \cite{tomasello03}. Corpora make it possible to observe statistical properties in children speech: size of the  vocabulary used, average length of the utterances, constructions effectively used, etc.  \cite{poibeau13}. Some other variables can be observed like the influence of adult speech on children production as well as social and gender influence. Corpora make observation more direct but do not provide a direct answer to more theoretical questions like the famous ``poverty of stimulus'' \cite{chomsky86,pullum02,legate02}. Lastly, researchers also tried to develop computational models that reproduce specific parts of the language acquisition process itself, like the acquisition of subcatgorization frames or of semantic categories \cite{macwhinney04,buttery07,alishahi08}. 

Most data included in Childes concern on average one hour of recording per month per child or more rarely per week, which means only a very small fractions of the child production is available. Not all the situations are represented so that mixing different corpora together does not always solve the problem: data may still be biased, quite unbalanced and unrepresentative. The project from Roy mentioned above \cite{roy09} was precisely supposed to overcome some of these limitations by providing a nearly exhaustive recording of the child input, but we have seen that this massive set of data is unstructured and therefore hard to process. 

\subsection{Language pathologies}

The study of the production of people with language pathologies has also attracted a high interest in the last decades, see for one example among many others \cite{haarmann1997aphasic}. This field can be compared, to a certain extent, to the research done in language acquisition: the idea is to get an accurate description of the production of people with languages pathologies so as to find what is deficient in their speech and then propose relevant treatments or relevant measures to help them overcome their difficulties \cite{grasemann11,kiran13}. Additionally cognitive science has of course a long tradition of mapping language deficiencies with specific areas in the brain. 

One of the main challenges for studies in language pathologies is to access large and representative corpora that are still lacking in the field \cite{ferguson08}. Most studies are based on small size data that may not be representative enough. Gathering large corpora is possible but pause multiple problems, from finding enough people with a similar pathology to ethical issues that are especially high here.  

\subsection{Language evolution}

At first sight, language evolution can be seen more as a social process than as a cognitive one. However, language evolution has to take into account how a group of individuals master a language and transmit this knowledge to their infants. This is the core of social cognition, that aims at studying how individual knowledge interacts so as to give birth to social processes. Language evolution can thus be seen as one of the central topics of social cognition \cite{seyfarth-cheney14,fitch10}.  It is our position here. 

Research on language evolution has been booming in the late 1990s and early 2000s giving birth to a series of important conference called Evolang (held every other year since 1996). Computational models play a major role in this field since they make it possible to observe the influence of various parameters in the evolution of languages. Christiansen and Kirby \cite{christiansen-kirby2003} propose to categorize research in this field in three broad categories:

\begin{enumerate}
\item \textit{evaluation}: computational models require to make explicit all the variables and parameters used and can thus be seen as a ``rigorous check''  of model hypotheses and help identify hidden variables; 
\item \textit{exploration}: simulations can be used to see how a population of agents evolve from a particular initial situation, materialized in the computational model through a particular settings of its parameters; 
\item \textit{exemplification}: computational models can be used as a tool to demonstrate ``how an explanation works'', especially for pedagogical purposes. 
\end{enumerate}
 
In fact, these different categories are not exclusive, and nearly all simulations serve at the same time as a proof, as an illustration, and provide new ideas for further experiments. Most simulations are based on multi-agent systems, where large populations of agents give birth to new generations of agents at regular intervals. They transmit their language to new generations according to various parameters as said above. The evolution of different linguistic features (addressing phonology, morphology, syntax or even semantics) can thus be studied in this framework, see for example \cite{briscoe09,oudeyer13,kirby14} among many other references. 

It should finally be noted that most simulations with multi-agent systems offer a quite abstract representation of the evolution of real languages (these simulations can even be completely abstract and related to populations of robots for example \cite{steels05}). This is probably one of the main reasons of the apparent recent decrease of interest towards this field of research: computational models proposed so far were quite abstract and are hard to prove or justify against real world data. One of the challenge is thus now to better connect these models with real world data, which is a long term and highly challenging task (see \cite{trijp12} for a recent example).

\section{Content of this special issue}

This special issue is dedicated to get a better picture of the relationships between computational linguistics and cognitive science. It specifically raises two questions: ``what is the potential contribution of computational language modeling to cognitive science?'' and conversely: ``what is the influence of cognitive science in contemporary computational linguistics?''

The call for papers specifically targeted  contributions on actual applications of methods from computational linguistics aiming at modeling cognitively motivated  phenomena, as well as applications of cognitive theories to the computational modeling of language. 

The following topics were proposed:

\begin{itemize}
\item Computational models of natural language acquisition, word clustering and word segmentation 
\item   Psycholinguistically motivated phonetic, phonological, morphological syntactic, semantic, pragmatic studies of language 
\item   Statistical and probabilistic modeling of factors encouraging one production or interpretation over its competitors 
\item   Models of language emergence, change and evolution 
\item   Models of language processing and surprisal 
\item   Experimental or corpus driven modeling and analysis of language
\end{itemize}

We have received seven submissions, and four of them have been selected for publication (two of them written in English and two in French). We think they give a good overview of some recent research trends: the four articles cover very different areas of the field, although they of course do not represent the whole field. 

The first paper, by Maxime Amblard, Kar\"en Fort, Caroline Demily, Nicolas Franck and Michel Musiol is entitled ``Analyse lexicale outill\'ee de la parole transcrite de patients schizophr\`enes'' (``Machine-assisted lexical analysis of speech transcripts from schizophrenic patients''). The paper reports some recent results obtained from the analysis of transcriptions of audio recordings of  people suffering from schizophrenia. The corpus at stake contains 375.000 tokens, which is considerably larger than previous similar corpora and makes it necessary to use natural language processing tools for the analysis. The paper mainly investigates disfluencies and the use of lexical forms. The main conclusion is that although people with schizophrenia produce more disfluencies, their discourse does not seem specific concerning the use of lexical forms, contrary to a commonly held hypothesis. This paper illustrates how natural language processing now plays a major role in understanding language pathologies. 

The second paper, by Quentin Feltgen, Benjamin Fagard and Jean-Pierre Nadal is entitled ``Repr\'esentation et \'evolution du langage dans les mod\`eles num\'eriques : la grammaticalisation comme perspective'' (``Language representation and models of language evolution: A grammaticalization perspective''). The authors investigate the notion of ``grammaticalization'', a well known linguistic phenomenon where some words or expressions become gradually frozen so as to play a grammatical role (the most well known example in French is the noun ``\textit{pas}'' that has completely lost its original meaning when the word is used as a negation in contemporary French). Even if a large number of recent projects in computational linguistics have explored language evolution, phenomena such as grammaticalization seem to have been put aside in the computational community, despite the high success of this notion among historical linguists. Within this context, the authors propose a new model illustrating, among other things, the loss of semantic content, known as ``semantic bleaching'' \cite{givon79} (often translated as ``javellisation s\'emantique'' in French). 

The third paper, by \'Akos K\'ad\'ar, Afra Alishahi and Grzegorz Chrupa{\l}a is entitled ``Learning word meanings from images of natural scenes''. The paper investigates how to associate word meanings with objects (through images) in noisy environments. To address this question, the authors use a large collection of images with natural language descriptions. The idea is to detect regularities and gradually create associations between words and image features. The authors show that their model correlates with human similarity judgments of word pairs when taking into account ambiguity and referential uncertainty. A parallel can be drawn with learning words in real life environments where children need to identify which aspects of the scenes are related to which parts of the perceived utterances. 

The last paper, from Bruno Gaume, Karine Duvignau, Emmanuel Navarro, Yann Desalle, Hintat Cheung, Shu-Kai Hsieh, Pierre Magistry and  Laurent Pr\'evot is entitled ``Skillex: a graph-based lexical score for measuring the semantic efficiency of used verbs by human subjects describing actions''. The authors are interested in the conceptual organization of lexical networks. They describe a technique to derive a semantic network from a dictionary and show that, despite surface disagreements, networks derived from different dictionaries all share the same topological structure. They assume that this structure is meaningful from a cognitive point of view and they show how this assumption can be used to evaluate action labelling tasks. They present a technique for the evaluation of what they call ``naming efficiency'' and detail a comparison involving children and adults.

\section*{Acknowledgements}

We thank the journal editors \'Eric de la Clergerie, Yves Lepage,
Jean-Luc Minel and Pascale S\'ebillot, as well as the following reviewers, for their 
efforts in making this special issue possible:  

\begin{itemize}
\item F. Alario (LPC, Univ. Aix-Marseille, France), 
\item A. Alishahi (Tilburg University, The Netherlands), 
\item M. Amblard (LORIA, Univ. de Lorraine, France), 
\item P. Blache (LPL, Aix-en-Provence, France),  
\item A. Christophe (LSCP, Paris, France), 
\item S. Colonna (SFL, Univ. Paris-VIII, France),  
\item I. Dautriche (LSCP, Paris, France), 
\item E. Dupoux (LSCP, Paris, France), 
\item T. Francois (Cental, UCL, Belgium),  
\item B. Gaume (CLLE-ERSS, Toulouse, France),  
\item J. Ginzburg (CLILLAC, Univ. Paris-VII, France),  
\item M. Grant (McGill University, Canada), 
\item J. Hale (Cornell University, USA), 
\item B. Hemforth (LLF, Univ. Paris-VII, France),  
\item P. Logacev (Potsdam University, Germany), 
\item F. Pellegrino (DDL, Lyon, France), 
\item D. Reitter (Penn State University, USA),  
\item E. Shutova (U. Berkeley, USA), 
\item J. Thuilier (CLLE-ERSS, Univ. Toulouse Jean-Jaur\`es, France), 
\item A. Villavicencio (UFRGS, Brazil), 
\item T. von der Malsburg (UC San Diego, USA), 
\item M. Zock (LIF, Marseille, France).
\end{itemize}

\bibliography{TAL_biblio_ex}

\end{document}